\documentclass[11pt]{article}
\usepackage{coling2020}
\usepackage{times}
\usepackage{url}
\usepackage{latexsym}
\usepackage{xcolor}
\usepackage{graphicx,url}
\usepackage{enumitem}

\title{NILC-Metrix: assessing the complexity of written and spoken language in Brazilian Portuguese}

\author{
  \AND \hspace{-1cm} Sidney Evaldo Leal \textsuperscript{1} \\
  \hspace{-1cm}{\small{\tt sidleal@gmail.com}} \\\And
  \hspace{-1cm}  Magali Sanches Duran \textsuperscript{1} \\
  \hspace{-1cm}{\small{\tt magali.duran@gmail.com} }  \\\And
  Carolina Evaristo Scarton \textsuperscript{2} \\
  {\small{\tt c.scarton@sheffield.ac.uk}} \\\AND
  Nathan Siegle Hartmann \textsuperscript{3}\\
  {\small{\tt nathanshartmann@gmail.com }} \\\And
  Sandra Maria Aluísio \textsuperscript{1}\\
  {\small{\tt sandra@icmc.usp.br}} \AND
  \\
  {\small \textsuperscript{1} Instituto de Ciências Matemáticas e de Computação - University of S\~{a}o Paulo, S\~{a}o Paulo, Brazil} 
  \\
 {\small \textsuperscript{2} The University of Sheffield, Sheffield, UK}
 \\
 {\small \textsuperscript{3} Itaú Unibanco, São Paulo, Brazil} \\
}

\date{}

\begin{document}
\maketitle
 \vspace{80pt}
\begin{abstract}
This paper presents and makes publicly available the NILC-Metrix, a computational system comprising 200 metrics proposed in studies on discourse, psycholinguistics, cognitive and computational linguistics, to assess textual complexity in Brazilian Portuguese (BP). These metrics are relevant for descriptive analysis and the creation of computational models and can be used to extract information from various linguistic levels of written and spoken language. The metrics in NILC-Metrix were developed during the last 13 years, starting in 2008 with Coh-Metrix-Port, a tool developed within the scope of the PorSimples project. Coh-Metrix-Port adapted some metrics to BP from the Coh-Metrix tool that computes metrics related to cohesion and coherence of texts in English. After the end of PorSimples in 2010, new metrics were added to the initial 48 metrics of Coh-Metrix-Port. Given the large number of metrics, we present them following an organisation similar to the metrics of Coh-Metrix v3.0 to facilitate comparisons made with metrics in Portuguese and English. In this paper, we illustrate the potential of NILC-Metrix by presenting three applications: (i) a descriptive analysis of the differences between children's film subtitles and texts written for Elementary School I\footnote{Comprises classes from 1st to 5th grade.} and II (Final Years)\footnote{Comprises classes from 6th to 9th grade, in an age group that corresponds to the transition between childhood and adolescence.}; (ii) a new predictor of textual complexity for the corpus of original and simplified texts of the PorSimples project; (iii) a complexity prediction model for school grades, using transcripts of children's story narratives told by teenagers. For each application, we evaluate which groups of metrics are more discriminative, showing their contribution for each task.
\end{abstract}

% \blfootnote{
%     \hspace{-0.65cm}  % space normally used by the marker
%     This work is licensed under a Creative Commons 
%     Attribution 4.0 International License.
%     License details:
%     \url{http://creativecommons.org/licenses/by/4.0/}.
% }

\section{Introduction}
\label{intro}

A set of metrics called NILC-Metrix was developed both in funded projects, involving multiple researchers, and in master's and doctoral projects at the Interinstitutional Center for Computational Linguistics --- NILC\footnote{\url{http://www.nilc.icmc.usp.br/}}, from 2008 to 2021. The motivation for developing this large set of metrics, the phases of its development, and also re-implementations of some metrics to make the use of Natural Language Processing (NLP) tools uniform, are summarised below. 

The initial motivation for building a set of metrics for automatic evaluation of textual complexity in BP started in the PorSimples project, whose theme was the Simplification of Portuguese Texts for Digital Inclusion and Accessibility \cite{candido-etal-2009-supporting,aluisio-gasperin-2010-fostering}. The target audience of PorSimples are people with low literacy, who want to obtain information from web texts but have some difficulty as they are literate at rudimentary and basic levels, according to the functional literacy indicator called INAF\footnote{\url{https://ipm.org.br/inaf}}.

In many projects of the reviewed literature, automatic text simplification is implemented as a process that reduces the lexical and/or syntactic complexity of a text while trying to preserve its meaning and information \cite{Carroll98practicalsimplification,Max2006,Shardlow2014}. However, there are simplification projects, for example, the Terence project, in which the target audience also requires simplifications to improve the understanding of the text both at the local level, helping to establish connections between close sentences and also at the global level of the text, helping in the construction of a mental representation of the text \cite{terence_2014}. There are still other initiatives, such as the Newsela\footnote{\url{https://newsela.com/}} company, which perform the conceptual simplification, simplifying the content, in addition to the form \cite{xu-etal-2015-problems}. Newsela also includes elaborations in the text to make certain concepts more explicit or the use of redundancies to emphasise important parts of the text. In addition, operations reduce and omit information that is not suitable for a given target audience. Based on the aforementioned simplification projects, we realised that textual complexity and textual simplification are strongly associated in the NLP area. We also realised that the type of simplification used in the Terence project aims to improve the coherence of a text, which makes the authors characterise this type of simplification as being at the cognitive level. The simplification done by Newsela is the most complete in terms of different operations, although still without complete automation (but see the advances carried out by \cite{alva-manchego-etal-2017-learning}).

During the project PorSimples, we implemented a system called \textbf{Facilita} responsible for adapting web content for low-literacy readers by using lexical elaboration and named entity labeling \cite{Watanabe2010}, and the simplification system was called \textbf{Simplifica}. One of Simplifica's particularities was to carry out two levels of simplification, called natural and strong, to help people who are literate at basic and rudimentary levels, respectively. To analyse the textual complexity of the resulting text, and thus assess whether the simplification goal had been achieved, a multiclass predictor of textual complexity was built using traditional machine learning methods. This predictor required the extraction of a set of metrics that could assess the complexity of a text and compute proxies to assess the cohesion and coherence of the simplifications supported by Simplifica's automatic rules. In this scenario, the Coh-Metrix-Port \cite{Scarton2010a,Scarton2010b,Scarton2010c} project was created. 

At NILC, we had already carried out a readability study before PorSimples aiming to adapt the Flesch Index to BP \cite{fleschport1996}, based on a corpus created to help identify the weights of the linear formula that evaluates word size and size of sentences in texts of various text genres and sources. The Flesch Index \cite{flesch1948} is based on the theory that the shorter the words and sentences are, the easier a text is to be read. Although it is very practical, as it is a number indicative of the complexity of the text and can be associated with school grades, it does not inform which operations to perform in a given text to reach the sizes of short sentences, for example. In addition, it can lead us to make mistakes, because a short text is not the only characteristic of an easy-to-read text. One of the criticisms of the Flesch Index and other traditional readability formulas \cite{dalechall1948,gunning1952,fry1968,kincaidetal1975} is that they are often used to adapt instructional material as prescriptive guides and not as simple predictive tools for textual complexity \cite{crossley2008assessing}. These mistakes derive from the failure to understand that the traditional readability formulas were not made to explain the reason for the difficulty of a text, as they are not based on theories of text understanding. Instead these formulas were based on the statistical correlation of superficial measures of a text with its level of complexity, previously established by a linguist or specialist in education, for example.

Once the limits of traditional readability formulas at the beginning of the Coh-Metrix-Port project were understood, we chose the Coh-Metrix project as a foundation for the metrics to be developed in PorSimples. Coh-Metrix computes computational cohesion and coherence metrics for written and spoken texts \cite{Graesser2004,Coh-Metrix_2011,graesser2014coh} based on models of textual understanding and cognitive models of reading \cite{kintsch1978toward,kintsch1973reading,kintsch1998comprehension} that explain:
(i) how a reader interacts with a text,
(ii) what types of memories are involved in reading, e.g., how the overload of working memory caused by using too many words before the main verb negatively influences the processing of sentences,
(iii) the role of the propositional content of the speech \cite{kintsch1998comprehension} which means that if the coherence of a text is improved, so will its comprehension \cite{crossley2007toward}, and
(iv) how the mechanisms of cohesion, for example, discourse markers and repetition of entities, will help to create a coherent text. 
In summary, just as the Coh-Metrix tool\footnote {\url{http://cohmetrix.com/}} for the English language does, the textual complexity analysis planned in Coh-Metrix-Port uses a framework of multilevel analysis. 
 
Coh-Metrix-Port provided 48 metrics grouped into 10 classes. %, described in detail in Section 2. 
However, one of its requirements was the use of open-source NLP tools. Thus, many syntactic metrics were not implemented, given the lack of free parsers with good performance at the time. Then, the AIC tool (Automatic Analysis of the Intelligibility of the Corpus) was created \cite{maziero2008ferramenta} within the scope of PorSimples. AIC has 39 metrics (most of them are syntactic) based on the parser Palavras \cite{bick2000palavras} (see details in Section 2).

After the end of PorSimples, in 2010, new metrics were added to the list of the initial 48 of the Coh-Metrix-Port tool and the 39 of the AIC. This was the case of the 25 new metrics of the Coh-Metrix-Dementia \cite{CBMS_CMD_2015,Aluisio_Propor16}, developed in a master's dissertation. During the implementation of Coh-Metrix-Dementia, the first re-implementation of Coh-Metrix-Port was done to standardise interfaces and the use of NLP tools. For example, the use of nlpnet PoS tagger  \cite{evaluating_2015} was set as the default tagger, as Coh-Metrix-Dementia incorporates the Coh-Metrix-Port's 48 metrics. In 2017, during a NILC student's PhD, a large lexical base with 26,874 words in BP was automatically annotated with concreteness, age of acquisition, imageability and subjective frequency (similar to familiarity) \cite{TSD2017}, enabling the implementation of 24 psycholinguistic metrics. 

The technology transfer project called \textit{Personalisation of Reading using Automatic Complexity Classification and Textual Adaptation tools} added 72 new metrics, many of them related to lexical and syntactic simplicity, to the already extensive set of metrics built by NILC.

Finally, the RastrOS project\footnote {\url{https://osf.io/9jxg3/?view_only=4f47843d12694f9faf4dd8fb23464ea9}} brought a new implementation to the 10 metrics based on semantic cohesion, via Latent Semantic Analysis (LSA) \cite{Landauer1997HowWC}, as well as for the calculation of lexical frequency metrics, now normalised. For the training of the LSA model with 300 dimensions, a large corpus of documents from the web, BrWaC \cite{wagner-filho-etal-2018-brwac}, was used. This same corpus was used, together with the corpus Brasileiro\footnote{\url{http://corpusbrasileiro.pucsp.br/}}, to calculate the lexical frequency metrics.

NILC-Metrix is, therefore, the result of various research projects developed at NILC. Its metrics were revised (some were rewritten, others discarded, several others had their NLP resources updated) and documented in detail between 2016 and 2017.  This documentation is available on the project's website. The metrics can be accessed via Web interface\footnote {\url{http://fw.nilc.icmc.usp.br:23380/nilcmetrix}} and its code is publicly available for download\footnote{\url{https://github.com/nilc-nlp/nilcmetrix}}, with an AGPLv3 license. Two of the parsers used by the metrics, Palavras and LX-parser \cite{LXparser_SilvaBCR10}, need to be installed, for the correct functioning of the metrics that depend on them; Palavras is a proprietary parser; LX-parser has a license that does not allow the parser to be distributed\footnote{\url{http://lxcenter.di.fc.ul.pt/tools/en/conteudo/LX-Parser_License.pdf}}.

In this paper, we present NILC-Metrix in detail and illustrate the potential of the tool with three applications of its metrics: (i) an evaluation of texts heard and read by children, showing the differences between the subtitles of films and children's series of the Leg2Kids project\footnote{\url{http://www.nilc.icmc.usp.br/leg2kids/}} and informative texts written for children in Elementary School I and II, compiled during the Coh-Metrix-Port and Adapt2Kids project \cite{Hartmann_Aluisio_2020}; (ii) a new predictor of textual complexity for the corpus of original and simplified texts of the PorSimples project, comparing its results with the predictor developed in \cite{aluisio-etal-2010-readability}; and (iii) a predictor of textual complexity, using narrative transcripts from the Adole-Sendo project\footnote{\url{adole-sendo.info/}}.

The remainder of this paper is organised as follows. Section 2 describes two tools developed during PorSimples that provided the basis for NILC-Metrix. Section 3 	presents the metrics, grouped into 14 classes, which is very similar to the organisation of the metrics used by Coh-Metrix v3.0, to make the comparative studies easier. Section 4 presents the corpora used in the NILC-Metrix applications and also the results of the three experiments with the metrics. Section 5 carries out a review analysing studies that used sets of metrics available in NILC-Metrix, in several research areas --- Natural Language Processing, Neuropsychological Language Tests, Education, Language and Eye-tracking studies. Finally, Section 6 presents some concluding remarks and suggests future work.

\section{Background: Coh-Metrix-Port and AIC tool Metrics}

In	 this	section, we present details of the two tools developed in the PorSimples project: Coh-Metrix-Port and AIC.	 The Coh-Metrix-Port provided 48 metrics grouped into 10 classes, shown below with the NLP tools and resources used in their implementation: 

\begin{enumerate}
\item Basic Counts contains 14 metrics related to basic statistics (average number of words per sentence and per paragraph, average number of letters per word, number of words and sentences in the text, and average number of syllables per content word), Flesch Index, and PoS related counts, using a model trained with the MXPOST tagger and the Nilc tagset\footnote{\url{http://www.nilc.icmc.usp.br/nilc/tools/nilctaggers.html}}; 
\item Logic operators contains 5 metrics related to the counting of logical operators AND, OR, IF, Negation;
\item Content word frequencies contains 2 metrics that use the largest lexicon that existed at the beginning of PorSimples, the \textit{Banco do Português}\footnote{\url{https://www.pucsp.br/pesquisa-seleta-2011/projetos/047.php}}, with 700 million words. These two metrics have been maintained in the current version of NILC-Metrix, but new frequency metrics, using larger corpora, have also been included;
\item Hypernyms and Ambiguity bring a metric that calculates the average number of hypernyms per verbs in sentences using the BP Wordnet.Br v.1.0\footnote{\url{http://www.nilc.icmc.usp.br/wordnetbr/}} and 4 metrics that calculate the impact of the number of senses (calculated based on the Electronic Thesaurus of Portuguese TeP 2.0\footnote{\url{http://www.nilc.icmc.usp.br/tep2/}}) for content words (verbs, nouns, adjectives and adverbs);
\item Tokens groups 3 metrics of lexical richness and level of formality: the well-known Type-Token Ratio and two more related to personal pronouns in phrases and text, implemented using a partial parser to identify noun phrases;
\item Constituents deal with 3 metrics related to the workload in working memory, computing modifiers within noun phrases, the number of noun phrases and the number of words before main verbs;
\item Connectives brings 9 metrics related to discursive markers that help to explain the temporal, causal, additive and logical relationships in the text, implemented based on the work of \cite{PardoN06};
\item Coreferences and Anaphoras bring 7 metrics that address referential cohesion, implemented using the MXPOST tagger, a stemmer and the Unitex-PB dictionary\footnote{\url{http://www.nilc.icmc.usp.br/nilc/projects/unitex-BP/web/index.html}}.
\end{enumerate}

AIC has 39 metrics, implemented mainly with information extraction from the Palavras parser \cite{bick2000palavras} and grouped into 5 classes, which deal with Basic Counts, Syntactic Information on Clauses, Density of Morphosyntactic Categories, Personalisation, and Discourse Markers:

\begin{enumerate}
\item Basic Counts contains 6 metrics related to basic statistics on: number of characters, number of words and number of sentences in the text; average number of characters per word, average number of words per sentence, and number of simple words, based on the \cite{Biderman1998} children's dictionary;
\item Syntactic Information brings 13 metrics about clause information in sentences, mainly extracted from the parser Palavras \cite{bick2000palavras}, such as: number of sentences in the passive voice, mode and average number of clauses per sentence, number of clauses, number of sentences (separated by the number of its clauses), number of clauses that start with coordinating conjunctions, number of clauses that start with subordinating conjunctions, 
number of coordinating conjunctions, number of subordinating conjunctions, number of verbs in the gerund, participle, infinitive and all 3 together;
\item Density of Syntactic and Morphosyntactic Categories, extracted using the parser Palavras \cite{bick2000palavras}, contains 8 metrics: number of adverbs, number of adjectives, number of prepositional objects and their average by clause and sentence, number of relative clauses, number of appositive clauses, number of adverbial adjuncts;
\item Personalisation contains 10 metrics related to the number of personal and possessive pronouns and their division by person and number;
\item Discourse Markers contains two metrics related to discursive markers, based on the work of \cite{PardoN06}: number of discursive markers and number of ambiguous discursive markers in the text. The latter are those that indicate more than one discourse relation. For example, in English “since” can function as either a temporal or causal connective.
\end{enumerate}

\section{NILC-Metrix Presentation}
\label{sec:nilcmetrix}

NILC-Metrix gathers 200 metrics developed over more than a decade for Brazilian Portuguese. The main objective of these metrics is to provide proxies to assess cohesion, coherence and textual complexity. Among other uses, NILC-Metrix may help researchers to investigate: (i) how text characteristics correlate with reading comprehension; (ii) which are the most challenging characteristics of a given text, that is, which characteristics make a text or corpus more complex; (iii) which texts have the most adequate characteristics to develop target learners’ skills; and (iv) which parts of a text are disproportionately complex and should be simplified to meet a given audience.
We hope that making the metrics available will stimulate new applications to validate them.
For the sake of presentation, the metrics are grouped into 14 categories, following their similarity and theoretical grounds. They are: Descriptive Index, Text Easability Metrics, Referential Cohesion, LSA-Semantic Cohesion, Lexical Diversity, Connectives, Temporal Lexicon, Syntactic Complexity, Syntactic Pattern Density, Semantic Word Information, Morphosyntactic Word Information, Word Frequency, Psycholinguistic Measures and Readability Formulas. 

\subsection{Descriptive Index}
Under this category we grouped the metrics that describe basic text statistics: number of words in the text; number of paragraphs in the text; number of sentences in the text; mean number of sentences per paragraph; mean number of syllables per content word; mean number of words per sentence; maximum number of words per sentence; minimum number of words per sentence; standard deviation of number of words per sentence; proportion of subtitles in relation to the number of sentences in the text.
The length of words, sentences and paragraphs correlates with the effort required to read a text. The standard deviation of words per sentence, as well as the maximum and minimum number of words per sentence, indicate how homogeneous a text is under this parameter. A large standard deviation is suggestive of large variations in terms of the number of words per sentence. If a text has many subtitles, this may affect the standard deviation.
These metrics do not require sophisticated resources to be processed: it is sufficient to have a tokeniser and sentence segmentation that recognise tokens, sentences and paragraph boundaries. 

\subsection{Text Easability Metrics}
This category brings together the metrics that measure how easy a text is. There are four measures that calculate the proportion of short, medium, long and very long sentences in relation to all sentences in the text (the four add up to 100\%). The classification of sentences according to their length is based on the following parameters: up to 11 words = short; between 11 and 12 = medium; between 12 and 15 = long; above 15 = very long.
Two other metrics of text easability accounts for the proportion of easy and difficult conjunctions to total words. The classification of conjunctions according to their easability is based on an informed lexicon.
Another metric of text easability is the proportion of first-person personal pronouns in relation to all personal pronouns in the texts. First-person personal pronouns indicate proximity to the reader.
Finally, the dictionary of Simple Words by \cite{Biderman1998} and a list of 909 concrete words from  \cite{JANCZURA2007} provided the lexicon used to calculate the proportion of simple content words to all content words in the text. Content words (nouns, verbs, adjectives and adverbs) constitute the variable vocabulary a reader has to know to understand the text (they oppose to function words, such as determiners, conjunctions, prepositions, numbers and pronouns, which do not point to extra linguistic referents). The greater the proportion, the simpler the text.

\subsection{Referential Cohesion}
There are nine metrics in this category and they capture the presence of elements necessary to construct coreference chains. These metrics calculate the overlap of content words in adjacent sentences and among all sentences of the text. Stem overlap is also calculated (such as in abolish-abolition). The longer the text, the greater the need of coreference chains to help the reader to make connections between parts of the text, rendering the text easier to understand.

\subsection{LSA-Semantic Cohesion}
The metrics that calculate semantic cohesion are grounded in Latent Semantic Analysis (LSA)\footnote{\url{http://lsa.colorado.edu/}} \cite{Landauer1997HowWC}, which considers the overlap of semantically related words. Co-occurrence is the basis to capture semantic relations. LSA uses Singular Value Decomposition (SVD) to reduce the complex matrix of words co-occurrences in a document to approximately 100-500 functional dimensions. Therefore, by representing the similarity of words in a vector space and computing the cosine of the angle between vectors of pairs of words, one can represent greater similarity with high cosines.
The LSA model for NILC-Metrix was trained on BrWaC\footnote{ \url{https://www.inf.ufrgs.br/pln/wiki/index.php?title=BrWaC}}, with 300 dimensions.  BrWac is the largest Brazilian corpus publicly available today (53 million documents, 2.68 billion words, and 5.79 million unique forms).

NILC-Metrix has eleven metrics of semantic cohesion. Six of them calculate the mean and the standard deviation of semantic overlap between: adjacent sentences, adjacent paragraphs and all sentence pairs in the text. The language model is also used to calculate the mean and the standard deviation of givenness (previous given information) and span \cite{Span_LSA_2003} (an alternative and better method to capture given information) in the current sentence. Finally, the cross-entropy calculates the mean difference of the probability distribution of sentence pairs in the language model.

\subsection{Psycholinguistic Measures}
NILC-Metrix brings six indices for each of the following psycholinguistic measures: age of acquisition, concreteness, familiarity and imageability, totalling 24 metrics. These measures are related to text easability: the lower the words’ age of acquisition, the easier the text, and the higher the words’ concreteness, familiarity and imageability, the easier the text. The lexical resource used by these metrics \cite{TSD2017} contains 26,874 words (content words), therefore if a word of the text is not included in the resource, these metrics are affected.

\subsection{Lexical Diversity}
Lexical diversity is a measure obtained through the type-token ratio (TTR), that is, the number of types (all words, disregarding repetitions) divided by the number of tokens (all words, considering repetitions). Lexical diversity is inversely proportional to cohesion: the lower the lexical diversity, the higher the cohesion. As explained by \cite{mcnamara_graesser_mccarthy_cai_2014}  \textit{TTR is correlated with text length because as the number of word tokens increases, there is a lower likelihood of those words being
unique}. NILC-Metrix includes TTR for: all words, content words, function words, nouns, verbs, adjectives, pronouns, indefinite pronouns, relative pronouns, prepositions and punctuation. Again, the detailed metrics are intended to investigate where the difficulty of the text lies.

\subsection{Connectives}
Connectives are words that help the reader to establish cohesive links between parts of the text.
NILC-Metrix provides metrics for the proportion of all connectives in the text, as well as for the proportion of four different types of connectives: additive, causal, logical and temporal. Temporal connectives, however, are within the temporal lexicon category. For each type, there is a distinct metric specifying the positive and negative ones. Besides that, the most frequent connectives, “e” (and), “ou” (or) and “se” (if) are focused on specific metrics.

\subsection{Temporal Lexicon}
The eleven indices gathered in this item detail the relative occurrences of each verb tense and mood in relation to the total verb tenses and moods in the text. Temporal connectives, positives and negatives, are also included in this category. The temporal lexicon is the first step towards enabling the construction of temporal cohesion metrics.

\subsection{Syntactic Complexity}
NILC-Metrix contains a series of metrics, using both dependency and constituency parsers. Some of them focus on syntax characteristics associated to the demand on working memory, as the number of words before the main verb. 
Using data from dependence trees, there are three metrics: distance in the dependency tree and two syntactic complexity indexes: Yngve \cite{Yngve1960} and Frazier \cite{Frazier1985}. Yngve’s index is based on the premise that the syntactic trees tend to branch to the right, and that deviations from this pattern correspond to greater complexity in the language. 

Frazier proposed a bottom-up approach, starting from the word and moving up the syntactic tree until it finds a node that is not the leftmost child of its parent. Each node in the tree receives a score of 1, and the nodes that are children of nodes of sentence type, 1.5. The score of each word is given by the sum of the scores of the nodes belonging to its branch.

In addition, the category of syntactic complexity brings various proportion measures involving clauses, enabling an in-depth investigation on where the complexity of a text lies: clauses with postponed subject; clauses in non-canonical order (canonical order is SVO: subject-verb-object), clauses in passive voice, infinite verb clauses, subordinate clauses, relative clauses, adverbial clauses, etc.

\subsection{Syntactic Pattern Density}
In this category, there are four metrics correlated with text processing difficulty: gerund clauses, mean number of words per noun phrase, maximum and minimum number of words per noun phrase. 

\subsection{Morphosyntactic Word Information}
In this category, one can find the traditional measures of content and functional word densities, in the text and per sentence, as well as a series of break-downs of these densities: adjectives, adverbs, verbs (inflected and non-inflected), nouns, prepositions, pronouns (detailed by type and inflection). Altogether there are 42 metrics that, although they do not individually give a measure of complexity, may be useful to investigate in detail where the difficulty of a text lies.

\subsection{Semantic Word Information}
This category has eleven metrics. Two of them use Brazilian Portuguese LIWC 2007 Dictionary\footnote{\url{http://143.107.183.175:21380/portlex/index.php/en/liwc}} to calculate the proportion of words with negative/positive polarity in relation to all words in the text. Five measures of ambiguity (of content words, and in detail by nouns, adjectives, verbs and adverbs) are calculated according to their respective number of senses in TeP (Portuguese Electronic Thesaurus). The average amount of hypernyms per verb in sentences uses information extracted from Wordnet.Br. Finally, there are three metrics relating to the proportion of abstract nouns and proper nouns in sentences and in the text.

\subsection{Word Frequency}
This category presents ten frequency measures. The two oldest present frequencies (not normalised) of all content words and of the rarer words in the text. They were extracted from Corpus do Português, which was the largest corpus at that time, with 700 thousand words. 
More recently, four frequency measures were extracted from Corpus Brasileiro \cite{Sardinha2004}, which has around one billion tokens 
and four from BrWaC, which has around 2.68 billion tokens \cite{wagner-filho-etal-2018-brwac}. The four measures are the same for the two corpora: average frequency of content words and rare content words; average frequency of all words and all rare words. The resulting eight measures were first normalised using fpm (frequency per million) and then normalised using the zipf logarithm scale. The difference between the two corpora is that Corpus Brasileiro assigned the PoS tags to the words out of context and for BrWaC, we assigned the PoS tags in context.

\subsection{Readability Formulas}
This category gathers five classic formulas used to assess text readability: 

The Brunet readability index \cite{Thomas_2005} is a kind of type/token ratio that is less sensitive to the text length. It raises the number of types to the constant -0.165 and then raise the number of tokens to the result.

The Dale Chall adapted formula \cite{dalechall1948} combines the percentage of unfamiliar words with the average number of words per sentence. Unfamiliar words are those not included in the Dictionary of Simple Words \cite{Biderman1998}. The calculus is: (0.1579 * percentage of unfamiliar words) + (0.0496 * average amount of words per sentence) + 3.6365.

The Flesch readability index \cite{kincaidetal1975} looks for a correlation between average word and sentence lengths. The formula after adaption is: 248.835 - [1.015 x (average words per sentence)] - [84.6 x (average syllables per word)].

Gunning’s Fog index\footnote{\url{https://core.ac.uk/reader/77238827}} adds the average sentence length to the percentage of difficult words and multiplies this by 0.4. Difficult words are those with more than two syllables. The result is directly related to the 12 American grade levels. 

Honore’s Statistics \cite{Thomas_2005} is a type/token ratio that takes into account, besides the number of types and tokens, the number of hapax legomena, that is, types that have only one token in the text.

\section{NILC-Metrix Applications}

In this section, we present three applications of NILC-Metrix metrics. Section 4.2 provides a comparison of texts heard and read by children, showing the differences between the legends of children's films and series from the Leg2Kids project and informational genre texts written for children in Elementary School I and II, compiled during the Coh-Metrix-Port and Adapt2Kids projects. Section 4.3 presents a new predictor of textual complexity for the corpus of original and simplified texts of the PorSimples project, comparing the results of the trained model with the 200 metrics of Nilc-Metrix with the predictor developed in \cite{aluisio-etal-2010-readability}, retrained with 38 metrics developed in the Coh-Metrix-Port project. Section 4.4 presents a predictor of textual complexity using transcripts of narratives from the Adole-sendo project to predict school grades. Section 4.1 describes the corpora used in the three experiments.

\subsection{Corpora used in the experiments}
\label{sec:sec4}

\subsubsection{Transcribed Legends of the Leg2Kids and Nonfiction Texts for Early School Years of the Adapt2kid projects}
\label{sec:leg_adapt_data}

The Leg2Kids corpus comprises 36,413 subtitles of films and a series of the genres Family and Animation in Brazilian Portuguese, made available by Open Subtitles\footnote{\url{https://www.opensubtitles.org}} in 2019.
The corpus was preprocessed to remove the existing time stamps in each subtitle (these markers define the time interval in which a subtitle will be displayed on the screen). Markings from the subtitle editors, such as web page addresses, acknowledgments, sponsorship, among others, were also removed. The corpus was then sentenced and tokenised by the NLTK\footnote{\url{https://www.nltk.org/}} tool.

Leg2Kids contains a total of 153,791,083 \textit{tokens} and 452,312 \textit{types}, and a \textit{type-token ratio} (TTR) of 0.29\%. This TTR value implies greater lexical richness than SUBTLEX-PT-BR \cite{tang2012Brazilian} (0.22\% TTR), a similar subtitle corpus in BP.

In order to build the Adapt2kids corpus for research on textual simplification for children \cite{Hartmann_Aluisio_2020},  we took advantage of some corpus already compiled in the PorSimples project, such as \textit{Ciência Hoje das Crianças} (CHC)\footnote{\url{http://chc.org.br/}}, \textit{Folhinha}\footnote{\url{http://www.folha.uol.com.br/folhinha}}, \textit{Para Seu Filho Ler}\footnote{\url{https://zh.clicrbs.com.br/rs}}.
To enlarge this corpus created during PorSimples, we selected the following sources: SARESP tests\footnote{\url{https://sites.google.com/site/provassaresp}} and textbooks for specific grades. SARESP tests are generally administered once a year; the test contains several textual genres – that is, there are few informative texts. We obtained only 72 texts, distributed in five grades, from SARESP tests. Regarding textbooks, we selected 178 informative texts from textbooks about the Portuguese language written in Portuguese. 
Because of the small amount of texts which had information about grade level, new sources were included in the corpus: NILC corpus\footnote{\url{http://nilc.icmc.usp.br/nilc/images/download/corpusNilc.zip}} and the magazine \textit{Mundo Estranho}\footnote{\url{http://mundoestranho.abril.com.br}}, which  contains 7,645 texts. 
The source distribution of Adapt2Kids corpus is shown in Table \ref{tab:corpusdistribution}.

\begin{table}[!htbp]
	\center
	\begin{tabular}{c|c|c|c|c|c|c}
      \hline
  & NILC & SARESP  & \textit{Ciência Hoje}& \textit{Folhinha}  & \textit{Para seu Filho Ler}  & \textit{Mundo} \\
		  Textbooks & corpus  & tests & \textit{das Crianças}   & Issue of Folha  & Issue  of & \textit{Estranho}\\
       & & & & de São Paulo &  Zero Hora & \\
      \hline
		492 & 262 & 72 & 2.589 & 308 & 166 & 3.756\\
      \hline
	\end{tabular}
	\caption{Distribution of Adapt2Kids texts by source.}
\label{tab:corpusdistribution}
\end{table}

From these 2 large corpora, we selected 2 samples with the same number of texts (see Table \ref{tab:corpora_Leg2Kids}) by: (i) selecting Adapt2Kids texts whose number of tokens is greater than 100, totalling 7,136 texts; (ii) selecting 7,136 texts of Leg2Kids longer than 600 tokens.
% Passei o comentário sobre o ESCOLEX para cá.
Leg2Kids has a \textit{type-token ratio} (TTR) of 0.29\%, but the sample selected of this corpus has a TTR of 0.012\%. The sample selected of Adapt2Kid has a TTR of 0.04\% implying greater lexical richness than Leg2Kids' sample but less lexical richness than Escolex \cite{Soares2014ESCOLEXAG} (1.5 \% TTR), which comprises 171 textbooks in European Portuguese for children attending the 1st to 6th grades (6- to 11-year old children) in the Portuguese education system. 
%Since the genre of Escolex texts is different to subtitles, one can expect a greater lexical diversity, given the dichotomy between written and spoken language.

\begin{table}[!htbp]
\centering
\begin{tabular}{llcccccc}
\hline
 \textbf{Corpus} & \textbf{Texts} & \textbf{Sent} & \textbf{ASL} & \textbf{Types} & \textbf{Tokens} & \textbf{TTR} \\ \hline
Leg2Kids & 7,136 & 2,170,971 & 6.18 & 148,004 & 11,972,556 &  0.012  \\
Adapt2Kids & 7,136 & 133,685 & 17.37 & 85,063 & 2,148,929 &  0.04  \\
\hline
\end{tabular}
\caption{Description of samples of Leg2Kids and Adapt2kids corpora}
\label{tab:corpora_Leg2Kids}
\end{table}

\subsubsection{Original and Simplified Texts of the PorSimples Project}

The PorSimples project has 154 original texts, considered complex for the target public, which were manually simplified on 2 levels, called natural simplification and strong simplification
(see Table ~\ref{tab:corpora_porsimples}). The result of the process is a parallel corpus with 462 texts. These two types of simplifications were proposed to attend the needs of people with different levels of literacy.

\begin{table}[!htbp]
\centering
\begin{tabular}{lcccccc}
\hline
\textbf{Level} & \textbf{Texts} & \textbf{Sent} & \textbf{ASL} & \textbf{Types} & \textbf{Tokens} & \textbf{TTR} \\ \hline
Original & 154 & 2960 & 19.99 & 11,106 & 57,237 &  0.19 \\ 
Natural & 154 & 4078 & 15.76 & 9,792 & 59,420 &  0.17 \\
Strong & 154 & 4918 & 12.76 & 9,647 & 60,760 &  0.16 \\ 
\textbf{Total} & 462 & 11,956 & 16.17 & 12,053 & 177,417 &  0.06 \\
\hline
\end{tabular}
\caption{Description of PorSimples corpus}
\label{tab:corpora_porsimples}
\end{table}

In PorSimples, the human annotator was free to choose which operations to use when performing a natural simplification, among the ones available, and when to use them. The annotator could decide not to simplify a sentence, for example. Strong simplification, on the other hand, was driven by explicit rules from a manual of syntactic simplification also developed in the project, which states when and how to apply the simplification operations. 

The simplifications were supported by an Annotation Editor \cite{Caseli2009}. The Annotation Editor has two modes to assist the human annotator: a Lexical and a Syntactical mode. In the Lexical mode, the editor proposes changes in words and discourse markers by simpler and/or more frequent ones, using two linguistic resources: (1) a list of simple words extracted from \cite{Biderman1998} and a list of concrete words from \cite{JANCZURA2007} and (2) a list of discourse markers extracted from the work developed by \cite{PardoN06}.  The Syntactical mode has 10 syntactic operations based on syntactic information provided by the parser Palavras \cite{bick2000palavras}. The syntactic operations, which are accessible via a pop-up menu, are the following: (1) non simplification; (2) simple or (3) strong rewriting; (4) putting the sentence in its canonical order (subject-verb-object); (5) putting the sentence in the active voice; (6) inverting the clause ordering; (7) splitting or (8) joining sentences; (9) dropping the sentence or (10) dropping parts of the sentence. 

\subsubsection{Transcribed Narratives of the Adole-sendo Project}

Adole-sendo is a project being developed at the Federal University of São Paulo (UNIFESP) that aims to assess biopsychosocial factors that affect the development of teenage (from 9 to 15 years old) behavior according to biological maturation measures. Here, we use only chronological age and related grades to train a complexity predictor of the narratives the teenagers produced setting a baseline for the Adole-sendo project. Currently, there are data collected from 271 participants, according to the distribution shown in Table \ref{tab:corpora3}.

\begin{table}[!htbp]
\centering
\begin{tabular}{lccccccc}
\hline
\textbf{Stages of Education} & \textbf{Grade} & \textbf{Texts} & \textbf{Sent} & \textbf{ASL} & \textbf{Types} & \textbf{Tokens} & \textbf{TTR} \\ \hline
Elementary   & 4th & 9 & 188 & 16.89 & 572 & 2,844 & 0.20 \\ 
School I     & 5th & 34 & 749 & 16.38 & 1,089 & 11,026 & 0.10 \\ \hline
Elementary   & 6th & 70 & 1,234 & 20.15 & 1,502 & 22,137 & 0.07 \\ 
School II    & 7th & 43 & 973 & 20.13 & 1,368 & 16,090 & 0.09 \\ 
             & 8th & 15 & 323 & 23.13 & 791 & 5,724 & 0.14 \\ 
             & 9th & 59 & 718 & 25.15 & 1,323 & 15,204 & 0.09 \\ \hline
High School  & 1st & 41 & 603 & 26.58 & 1,271 & 12,615 & 0.10 \\  
\textbf{Total}        &     & 271 & 4,788 & 21.80 & 3,129 & 85,640 & 0.04 \\
\hline
\end{tabular}
\caption{Description of Adole-sendo corpus}
\label{tab:corpora3}
\end{table}

The data for this project comprises transcribed narratives obtained from the task of telling children's stories from memory (referred to as retelling herein) for each adolescent. There are two stories used by participants in the collection: Jack and the Beanstalk and Little Red Riding Hood. The participant chooses one of the two.

The process of creating the corpus of the Adole-Sendo project included three steps: (i) transcription of the retelling audios and annotation of six linguistic phenomena at the word level; (ii) linguistic annotation of five types of disfluencies; (iii) automatic generation of narratives without the disfluencies and the phenomena annotated in the transcripts. These stages are summarised below.

\paragraph{Transcription of Children's Story Retelling Audios.}

The transcripts were obtained from retelling audios recorded during interviews with the participants at their own schools in the city of São Paulo, Brazil, (through convenience sampling), following the standard Portuguese spelling and sentence segmentation according to the rules of the language, based on prosodic and syntactic rules for written texts. 
The transcripts also included syntactic errors and filled pauses. Pronunciation variations were not transcribed.
Linguistic phenomena at the word level, such as intentional word repetition, three types of pauses, filled pauses and interrupted words were annotated during transcription.

\paragraph{Linguistic Annotation of Disfluencies.} 

Once the transcription stage was completed, 237 texts were imported into the web platform Inception\footnote{\url{https://inception-project.github.io/}} for annotating five phenomena of disfluencies, described in an annotation manual: (i) Discourse markers; (ii) Comments not related to the story were annotated; (iii) Repetitions of unintended words; (iv) Self-corrections; and (v) Filled pauses.

After a pilot annotation, three pairs of annotators each received a third of the transcriptions. The annotation comprises two tasks: (i) finding and correctly segmenting the phenomenon and (ii) correctly labeling the delimited phenomenon.
We used these two agreement measures as Cohen's Kappa only analyses chunks that the pair has segmented equally and excludes incomplete or missing annotations. The annotation obtained had high Kappas but the pairs had problems in finding the phenomena and segmenting them correctly. For this reason, we analysed the Alpha measure which considered the character-by-character annotation of the narrative, penalising cases in which the annotator exceeded one character (or more characters) or forgot to annotate a given phenomenon. In the annotation made, Alphas were negative for all pairs in the performed annotation, indicating that the annotators had difficulties in following the annotation manual or that they adopted different rules, consistently. The Alpha measure showed that the manual should be revised and agreed upon between the pairs of annotators to proceed with the curation stage to produce the gold standard annotation. The corpus was curated by the pair's most experienced annotator and the most difficult cases were discussed among all annotators.  

\paragraph{Automatic Generation of Narratives without Disfluencies.}

The linguistic annotations were computed to be used in the analysis of narratives, and a modified narrative –- without the six annotations taken during the transcription and without the five annotations of disfluencies –- was generated automatically. Afterwards, 34 new transcripts were annotated by an experienced annotator, totalling 271 narratives in the corpus.

\subsection{Comparison between spoken and written texts targeting children}
\label{sec:legvsadapt}

Written and spoken language are distinctive in nature. Whilst written texts are usually self-contained and well-structure, spoken language can use extra knowledge (information from pragmatics) in order to be unambiguous. The question we aim to answer in this section is whether or not we can quantify the differences between the two modalities by using metrics from NILC-Metrix. As in the original Coh-Metrix, NILC-Metrix's  metrics can capture different aspects of textual complexity that we will use to compare written (Adapt2Kids) and spoken (Leg2Kids) language. Our experiments are similar to the work of \cite{louwerse-etal-2004} that uses Coh-Metrix to compare written and spoken language in English. 

In order to perform the analysis, we use the Welch's t-test \cite{welch}: an extension of the t-test for distributions with unequal variances. We consider a statistically significant difference between the means of the distributions when the $p-$value is smaller than 0.001. Each metric is analysed in isolation and we discard metrics that cannot be applied to our texts (for example, paragraph-related metrics). Our analysis is divided into the 14 categories of NILC-Metrix. 

\subsubsection{Descriptive Index}

Although these are the most basic metrics in NILC-Metrix, they already provide some insight into the main differences between the two corpora. For instance, as expected, Leg2Kids subtitles have, on average, a significantly higher number of words (1,638.26) and sentences (292.06) than texts in Adapt2Kids (300.28 and 18.75, respectively). On the other hand, the word per sentence ratio is smaller in Leg2Kids (6.38) than in Adapt2Kids (17.17). In addition, the standard deviation of the sentence length is significantly higher in Adapt2Kids (9.57) than in Leg2Kids (5.11). This analysis highlights some of the main characteristics of Leg2Kids: subtitles consist of longer texts in terms of sentences, although they have less words per sentence (which is expected for dialogues, mainly subtitles with screen display constraints). Interestingly, there are no significant differences in terms of the maximum sentence length, i.e., on average, the longest sentences are similar in both corpora (around 37 words). Table \ref{tab:desc} shows the metrics with significantly higher values in Adapt2Kids than Leg2Kids (first row) and vice-versa (second row). 

\begin{table}[!htbp]
\centering
\begin{tabular}{lp{9cm}}
\hline
Adapt2Kids & syllables per content word, words per sentence, min sentence length, sentence length standard deviation  \\
\hline
Leg2Kids & number of sentences, number of words  \\
\hline
\end{tabular}
\caption{Descriptive Index: first line means higher values in Adapt2Kids and second line means higher values in Leg2Kids.}
\label{tab:desc}
\end{table}

\subsubsection{Text Easability Metrics}

All metrics from this category show significant differences. In particular, Leg2Kids (0.69) has a higher ratio of personal pronouns than Adapt2Kids (0.23), which is expected in the dialogue modality. Long sentences are more frequent in Adapt2Kids (ratio=0.51) than Leg2Kids (ratio=0.04), which is also a characteristic of subtitles. Finally, Leg2Kids has a higher ratio of simple words (0.76) than Adapt2Kids (0.74). Table \ref{tab:te} summarises the metrics with significantly higher values in Adapt2Kids than Leg2Kids (first row) and vice-versa (second row). 

\begin{table}[!htbp]
\centering
\begin{tabular}{lp{9cm}}
\hline
Adapt2Kids & ratio of easy(hard) conjunctions, ratio of very long(long/medium) sentences\\
\hline
Leg2Kids & ratio of personal pronouns, ratio of short sentences, ratio of simple words \\
\hline
\end{tabular}
\caption{Text Easability Metrics: first line means higher values in Adapt2Kids and second line means higher values in Leg2Kids.}
\label{tab:te}
\end{table}

\subsubsection{Referential Cohesion}

Three metrics in this category do not show statistically significant differences: proportion of adjacent references, argument overlap and mean of co-referent pronouns. For both corpora, the values of the metrics are low, suggesting they are both not complex. All other metrics show higher values in Adapt2Kids than Leg2Kids, indicating that written texts have more ambiguous pronouns, although they also have lexical repetition, which is a characteristic of simple texts. On the other hand, in dialogue, pronouns are usually easily solved, as most of them address the interlocutor/speaker. In addition, dialogue is dependent on extra-textual context and elements from pragmatics that may impact their intelligibility. Therefore, we cannot clearly conclude that written texts are simpler than subtitles. Instead, our analysis shows that written texts use more artifacts of referential cohesion than dialogues in subtitles.

\subsubsection{LSA-Semantic Cohesion}

In general, Adapt2Kids shows higher values of LSA-Semantic than Leg2Kids, which may suggest that the written texts present high semantic similarity among their sentences. This is not surprising, given that in Leg2Kids scenes they are not explicitly identified and, therefore, changes in topics may occur with higher frequency than in written texts. One exception is the LSA metric measured using the sentence span, where Leg2Kids shows, on average, a higher value (0.93) than Adapt2Kids (0.86). Differently from simply calculating the cosine similarity between a sentence vector and the average vector of its predecessors, in the span case the previous sentences are used to form a vector sub-space. The current sentence vector is decomposed in two components: one in the previous sentence sub-space and another perpendicular to this sub-space. The similarity score is then calculated between these two components and it is expected to measure similarity beyond the explicit content presented in previous sentences. This is an interesting result, suggesting that LSA semantic cohesion in subtitles needs to be measured using context beyond explicit clues. Table \ref{tab:lsa-sem} summarises the metrics with significantly higher values in Adapt2Kids than Leg2Kids (first row) and vice-versa (second row).

\begin{table}[!htbp]
\centering
\begin{tabular}{lp{9cm}}
\hline
Adapt2Kids & LSA adjacent (all/givenness) mean, LSA adjacent (all/givenness) standard deviation \\
\hline
Leg2Kids & LSA span mean, LSA span standard deviation, cross entropy \\
\hline
\end{tabular}
\caption{LSA-Semantic Cohesion: first line means higher values in Adapt2Kids and second line means higher values in Leg2Kids.}
\label{tab:lsa-sem}
\end{table}

\subsubsection{Psycholinguistic Measures}

\paragraph{Concreteness} Adapt2Kids has, on average, a higher concreteness score than Leg-2Kids (4.30 and 4.08, respectively). This happens mainly because Adapt2Kids has a high ratio of concreteness of words with scores between 4 and 5.5, whilst Leg2Kids has a high ratio of concreteness of words with scores between 2.5 and 4. Therefore, written texts in Adapt2Kids use significantly more concrete words than spoken language in Leg2Kids.

\paragraph{Familiarity} In terms of familiarity, Leg2Kids has a higher average score than Adapt2Kids (5.12 and 4.84, respectively). Leg2Kids shows a significantly higher ratio of words with familiarity scores between 5.5 and 7, whilst the highest ratio for Adapt2Kids for words with familiarity scores is between 4 and 5.5. Contrary to the concreteness results, Leg2Kids subtitles have significantly more familiar words than Adapt2Kids. 

\paragraph{Age of Acquisition} Adapt2Kids has a higher mean value of age of acquisition score (4.54) than Leg2Kids (3.72), suggesting that the subtitles are more accessible for younger ages. This happens because Leg2Kids has a high number of words with age of acquisition scores below 4, while most words in Adapt2Kids have scores higher than 4. 

\paragraph{Imageability} Leg2Kids and Adapt2Kids are not significantly different in terms of imageability. Although Leg2Kids shows a significantly higher value of words with scores between 4 and 5.5 (0.69) than Adapt2Kids (0.65); the absolute difference is rather small to draw any conclusions. Similarly, on the range of scores between 2.5 and 4, Adapt2Kids shows a significantly higher score than Leg2Kids, but the absolute difference is also very small (0.25 versus 0.23).

\subsubsection{Lexical Diversity} 

Most metrics in the lexical diversity category show significantly higher values in the Adapt2Kids texts. This indicates a higher complexity for written texts than subtitles. For instance, the type-token ratio scores for Adapt2Kids were 0.75, whilst Leg2Kids scored 0.74. In terms of content word diversity, Adapt2Kids also showed a higher score (0.84) than Leg2Kids (0.79). The exceptions where content density (that measures the proportion of content words in relation to functional words) and maximum proportion of content words (that shows the proportion of content words in the most complex sentence of a document). For these two metrics, Leg2Kids shows significantly higher values (1.74 for content density and 0.84 for maximum proportion content words) than Adapt2Kids (1.48 and 0.73 for content density and maximum proportion of content words, respectively).

\subsubsection{Connectives and Temporal Lexicon}

Except for the ratio of positive causal connectives and ratio of negative logical connectives, all other metrics showed statistically significant differences between both corpora. However, most values in this category are considerably small (all ratio values are below 0.1), suggesting that connectives are scarce in both written and spoken language. Adapt2Kids shows the highest ratio of all connectives (0.09 vs 0.08), whilst Leg2Kids has the highest ratio of negations (0.03 vs 0.01).

In the Temporal Lexicon category, the only metric that does not show statistically significant differences is the ratio of positive temporal connectives. Similar to connectives, most values are below 0.1 and Adapt2Kids has the highest values for the majority of metrics, indicating that written texts make more use of this type of connectives. Adapt2Kids also has the highest proportion of verbs in the present tense, suggesting that written texts have more frequent verb inflexions (0.57 vs. 0.22). On the other hand, Adapt2Kids has the highest proportion of auxiliary verbs followed by a verb in the past participle tense (0.14 vs. 0.01), which is a sign of higher complexity, but may also indicate that written texts are more formal than subtitles. Finally, Adapt2Kids also shows a higher proportion of different verb tenses (4.38) than Leg2Kids (3.61), which may also be capturing the characteristic of narrativity in subtitles, which 
implies in using the past tense frequently \cite{graesser2014coh}.

\subsubsection{Syntactic Complexity and Syntactic Pattern Density}

In general, Adapt2Kids shows higher syntactic complexity than Leg2Kids with a significant difference. 
The only exception is the metric measuring the distance in a parse tree, which did not show any statistically significant differences in the results. Table \ref{tab:syntactic} shows a selection of metrics in this category and their mean values for both Adapt2Kids and Leg2Kids (higher values mean higher syntactic complexity). These results highlight that subtitles (and dialogue in general) use simplified syntax. However, it is worth mentioning that, in Leg2Kids, sentences were automatically devised, as the subtitles were divided into the frames they appear in the screen. Therefore, more investigation is needed to draw further conclusions.

\begin{table}[!htbp]
\centering
\begin{tabular}{lcc}
\hline
 & Adapt2Kids & Leg2Kids \\
\hline
words before main verb & 1.51 & 0.80 \\
adverbs before main verb & 0.26 & 0.09\\
clauses per sentence & 2.35 & 0.46 \\
coordinate conjunctions per clauses & 0.04 & 0.23 \\
frazier & 7.06 & 5.99 \\
proportion of non-SVO clauses & 0.33 & 0.11\\
proportion of relative clauses & 0.13 & 0.02 \\
proportion of subordinate clauses & 0.44 & 0.11 \\
yngve & 2.48 & 1.60\\
\hline
\end{tabular}
\caption{Results for selected syntactic complexity metrics.}
\label{tab:syntactic}
\end{table}

Similarly to the results in the Syntactic Complexity category, Adapt2Kids also shows the highest values for Syntactic Pattern Density metrics. For instance, the mean size of noun phrases is significantly higher in Adapt2Kids (4.91) than in Leg2Kids (2.11). 

\subsubsection{Morphosyntactic Word Information, Semantic Word Information and Word Frequency}

All metrics in the Morphosyntactic Word Information category show statistically significant differences. Leg2Kids has the highest proportion of content words (0.62 vs. 0.59), while Adapt2Kids shows the highest proportion of functional words (0.41 vs. 0.38). Adapt2Kids has the highest noun (0.33 vs. 0.25) and adverb (0.77 vs. 0.37) ratios, whilst the ratio of pronouns (0.15 vs. 0.08) and verbs (0.24 vs. 0.16) are highest in Leg2Kids. Adapt2Kids also has the highest values for the proportion of infinitive verbs (0.18 vs. 0.07), inflected verbs (0.61 vs. 0.27) and non-inflected verbs (0.34 vs. 0.10). The ratio of prepositions per clause and per sentence is considerably higher in Adapt2Kids (1.35 and 2.73, respectively) than in Leg2Kids (0.17 and 0.21 respectively). The proportion of relative pronouns is also higher in Adapt2Kids (0.27) than in Leg2Kids (0.03). Finally, whilst the proportion of third person pronouns is the highest in Adapt2Kids (0.57 vs. 0.30), Leg2Kids shows the highest values for the proportions of second (0.32 vs. 0.2) and first person (0.37 vs. 0.05) pronouns. 

In the Semantic Word Information category, the only metric that does not show statistically significant differences is the proportion of negative words. Leg2Kids shows the highest values for metrics measuring the ambiguity of adjectives (5.01 vs. 3.60), nouns (2.49 vs. 2.29), verbs (10.95 vs. 9.75) and content words (6.17 vs. 4.47). The mean value of verb hypernyms and the proportion of positive words are higher in Adapt2Kids (0.56 and 0.39, respectively) than in Leg2Kids (0.38 and 0.34, respectively).

Finally, in the Word Frequency category, all metrics show statistically significant differences. The log of the mean frequency values for content words extracted from Corpus Brasileiro and BrWac are slightly higher in Leg2Kids (4.53 and 4.43, respectively) than in Adapt2Kids (4.51 and 4.28, respectively). When considering all words for the same metrics, Adapt2Kids shows slightly higher values than Leg2Kids. 

\subsubsection{Readability Formulas}

Table \ref{tab:readability} shows the average scores for each metric in this category for the different corpora (the differences are statistically significant). All metrics suggest that Leg2Kids is simpler than Adapt2Kids. However, it is worth emphasising that these readability metrics may not be capturing simplicity in our case. When analysing the Descriptive Indexes, we show that Leg2Kids has smaller sentences and smaller words than Adapt2Kids (words per sentence and syllables per content words metrics). Since readability metrics rely heavily on these two factors, it cannot be concluded that Leg2Kids is simpler than Adapt2Kids without any further analysis.

\begin{table}[!htbp]
\centering
\begin{tabular}{lcc}
\hline
 & Adapt2Kids & Leg2Kids \\
\hline
Brunet ($\uparrow$) & 11.03 & 12.87 \\
Adapted Dale-Chall ($\downarrow$) & 9.85 & 8.99\\
Flesch Reading Ease ($\uparrow$) & 51.72 & 76.35 \\
Gunning Fog ($\downarrow$)  & 7.00 & 2.65 \\
Honor\'e statistics ($\downarrow$)  & 1,040.01 & 933.04 \\
\hline
\end{tabular}
\caption{Results for readability metrics (arrows indicate the simplicity direction).}
\label{tab:readability}
\end{table}

\subsection{Complexity prediction of original and simplified texts using PorSimples corpus}

The PorSimples corpus of simplified texts was used to train a textual complexity model for the Simplifica   \cite{Scarton2010c} tool, which helped in the manual simplification process, supported by simplification rules. The model helps a professional to know when to stop the simplification process. In PorSimples, we had the mapping: natural - literate at a basic level; and strong - literate at a rudimentary level \cite{aluisio-etal-2010-readability}. The objective of the following experiment is to exemplify the use of NILC-Metrix metrics to classify these complexity levels.

In \cite{aluisio-etal-2010-readability}, the 42 Coh-Metrix-Port metrics are presented that are used for training a classifier for three levels of textual complexity. Here, we used 38 of these 42 metrics as four of them were discontinued due to a project decision in parser changing. The four discontinued metrics were: \textit{Incidence of NPs}, \textit{Number of NP modifiers}, \textit{Number of high level constituents} and \textit{Pronoun-NP ratio}.

Here, we try to answer two questions via machine learning experiments: (i) whether new features, described in Section~\ref{sec:nilcmetrix}, developed after the Coh-Metrix-Port project, add value to the task textual complexity prediction using the parallel corpus of PorSimples; and (ii) which categories of features best describe the characteristics that distinguish texts of the PorSimples project (original texts, naturally simplified and strongly simplified).

The method used was the Multinomial Logistic Regression, which has as its premise the ordinal relationship between classes (levels of simplification) \cite{heilman2008analysis}. This was the same method used in the original article of the Coh-Metrix-Port \cite{aluisio-etal-2010-readability} project. In order to better refine the analysis, we used the F1 metric by class and we also presented the F1 Macro, which provided us with a greater degree of detail regarding the difficulty of the task of classifying textual complexity. All experiments followed the stratified 10-fold cross-validation methodology when splitting the data between the training and testing sets. The stratified strategy ensures that all training and test folds contain all text levels, increasing the experiment's robustness. The division into 10 folds for training and testing is a good proxy for the leave-one-out methodology, ensuring good generalisation of the results achieved and greater confidence in a non-overfit or underfit result. We are aware of the small number of texts available for this experiment and the bias of such data volume analysis. Thus, it is essential to be careful about data usage.

\begin{table}[!htb]
    \centering
    \begin{tabular}{lrrrr}
        \hline
        Category &      Strong &      Natural &      Original & F1 Macro \\
        \hline
        All                                     &  0.655 &  \textbf{0.568} &  \textbf{0.888} &      \textbf{0.704} \\
        Coh-Metrix-Port & 0.719 & 0.514 & 0.806 & 0.679\\
        \hline
        Readability Formulas               &  \textbf{0.720} &  0.402 &  0.782 &      0.635 \\
        Syntactic Complexity                   &  0.675 &  0.409 &  0.813 &      0.632 \\
        Text Easability Metrics                  &  0.661 &  0.413 &  0.763 &      0.612 \\
        Morphosyntactic Word Information &  0.679 &  0.408 &  0.739 &      0.609 \\
        Descriptive Index                      &  0.701 &  0.284 &  0.734 &      0.573 \\
        LSA-Semantic Cohesion                         &  0.637 &  0.349 &  0.721 &      0.569 \\
        Lexical Diversity                      &  0.592 &  0.384 &  0.714 &      0.563 \\
        Referential Cohesion                       &  0.663 &  0.323 &  0.689 &      0.558 \\
        Semantic Word Information       &  0.422 &  0.331 &  0.671 &      0.475 \\
        Connectives                               &  0.506 &  0.286 &  0.623 &      0.472 \\
        Syntactic Pattern Density          &  0.577 &  0.269 &  0.551 &      0.466 \\
        Word Frequency                   &  0.477 &  0.318 &  0.582 &      0.459 \\
        Temporal lexicon                          &  0.552 &  0.232 &  0.530 &      0.438 \\
        Psycholinguistic Measures                &  0.394 &  0.250 &  0.593 &      0.412 \\
        \hline
        \end{tabular}
    \caption{Performance on PorSimples dataset. Results presented by category of features.}
    \label{tab:resultadosPorSimples}
\end{table}

Table~\ref{tab:resultadosPorSimples} presents the results of the automatic text classification experiment by the feature's category. This division gives us better visibility regarding the categories that most contribute to automatic classification, that is, those that best describe the characteristics that distinguish the original texts and their two levels of simplification. When comparing the use of all the features concerning the 38 of the Coh-Metrix-Port, we noticed it again in the macro F1 and also in the Natural and Original Classes, despite a slight worsening concerning the classification of the Strong class. Regarding feature categories, we noticed that the combination of all features presented the best F1 Macro for the task and also the best F1 micro for the Natural and Original classes. Regarding F1 for the \textit{Strong} class, we noticed that the individual use of the \textit{Readability Formulas} category presented a better result than its aggregated usage with other features. This result is interesting, as it presents us with a scenario in which the other groups of features confuse the classifier concerning the classification of this class. This confusion can occur due to the improvement in the distinction of the other classes (\textit{Natural} and \textit{Original}), causing a trade-off in relation to the \textit{Strong} class. In both evaluations, we noticed that the aggregate use of all features produces a slight worsening in the classification of the \textit{Strong} class, although it produces better results in general, which is positive in the end.

\begin{table}[!htbp]
    \centering
    \begin{tabular}{lr}
        \hline
        Category &  \# \\
        \hline
        Syntactic Complexity                   &     13 \\
        Word Frequency                   &      6 \\
        Descriptive Index                      &      5 \\
        Readability Formulas               &      5 \\
        LSA-Semantic Cohesion                         &      5 \\
        Lexical Diversity                      &      4 \\
        Text Easability Metrics                   &      3 \\
        Psycholinguistic Measures                &      3 \\
        Connectives                               &      2 \\
        Referential Cohesion                       &      2 \\
        Morphosyntactic Word Information &      2 \\
        Semantic Word Information       &      1 \\
        Syntactic Pattern Density          &      1 \\
        \hline
        \end{tabular}
    \caption{Features by category resulting from a Boruta procedure.}
    \label{tab:features_category_boruta}
\end{table}

We carried out a feature selection step to better understand which features are relevant in explaining the phenomenon of classification of PorSimples texts. We know that not all features are necessarily useful: some may not differentiate between simple and complex texts and others may be correlated with each other, that is, redundant. Therefore, we run the Boruta \cite{kursa2010boruta,kursa2010feature} method for feature selection. Boruta checks which features are more informative to explain the event of interest than a random variable produced from the shuffling of the feature itself. If a feature explains an event, it is correlated with the fact that a text is simple or complex, but if we scramble that feature, it loses its correlation with the event and no longer explains it. Boruta eliminated 147 of the 200 features, resulting in a subset of 53 features. Table~\ref{tab:features_category_boruta} shows the count of resulting features by category of features.

The justification for choosing Boruta among other selection methods was because the algorithm was designed to classify what the original article calls the “all relevant problem'': finding a subset of features that are relevant to a given classification task. This is different from the “minimum-optimal problem'', which is the problem of finding the minimum subset of features that perform in a model. Although the machine learning models in production should ultimately aim at selecting optimal minimum features, Boruta's thesis is that, for exploration purposes, minimal optimisation goes too far. Moreover, the method is robust to the correlation of features. In scenarios with a large number of features, dealing with their correlation can be a very costly task. Thus, using Boruta can also speed up the stage of preparing features, justifying our choice.

\begin{table}[!htb]
    \centering
    \begin{tabular}{lrrrr}
        \hline
        Category &      Strong &      Natural &      Original &  F1 Macro \\
        \hline
        All                                     &  0.708 &  0.508 &  \textbf{0.860} &     \textbf{0.692} \\
        Coh-Metrix-Port & 0.719 & \textbf{0.514} & 0.806 & 0.679\\
        \hline
        Readability Formulas               &  \textbf{0.720} &  0.402 &  0.782 &      0.635 \\
        Syntactic Complexity                   &  0.687 &  0.414 &  0.796 &      0.632 \\
        Text Easability Metrics                     &  0.644 &  0.389 &  0.752 &      0.595 \\
        Descriptive Index                      &  0.691 &  0.302 &  0.716 &      0.570 \\
       Morphosyntactic Word Information &  0.614 &  0.330 &  0.708 &      0.551 \\
        Lexical Diversity                      &  0.586 &  0.359 &  0.699 &      0.548 \\
        LSA-Semantic Cohesion                         &  0.590 &  0.295 &  0.672 &      0.519 \\
        Word Frequency                   &  0.468 &  0.321 &  0.600 &      0.463 \\
        Syntactic Pattern Density          &  0.557 &  0.271 &  0.554 &      0.461 \\
        Connectives                               &  0.500 &  0.219 &  0.555 &      0.425 \\
        Referential Cohesion                       &  0.307 &  0.340 &  0.540 &      0.396 \\
        Psycholinguistic Measures                &  0.410 &  0.227 &  0.531 &      0.389 \\
        Semantic Word Information       &  0.266 &  0.243 &  0.531 &      0.346 \\
        Temporal lexicon                          &  0.148 &  0.098 &  0.254 &      0.167 \\
        \hline
        \end{tabular}
    \caption{Performance on PorSimples dataset using only feature selected by Boruta. Results presented by category of features.}
    \label{tab:resultadosPorSimplesBoruta}
\end{table}

We replicated the PorSimples text classification experiment using only the features selected by Boruta. Table~\ref{tab:resultadosPorSimplesBoruta} presents the results obtained. Once more, we noticed that all feature usage (now the 53 selected ones) performed better in the classification of textual complexity concerning the 38 features replicated from Coh-Metrix-Port. We noticed a minimal difference in performance in the \textit{Strong} and \textit{Natural} classes but a significant gain in the \textit{Original} class, demonstrating value when using the new features. When comparing the use of the 53 selected features concerning the 200 features developed, we noticed a slight drop in the F1 Macro obtained, which can be justified by the small size of the dataset and weak correlations between the features, as well as between a feature and the target of the task. This kind of phenomenon tends to be irrelevant as the increase in the dataset causes effects such as these to be considered statistically insignificant. When we analyse the performance of the categories of features, the data show us that the difference in performance in the prediction of the \textit{Strong} class decreased between the use of all selected features and the use of the selected features of the \textit{Readability Formulas} category. While this difference tends to a rounding error, the combined performance of all the features selected in the prediction of the \textit{Natural} and \textit{Original} classes, as well as in the F1 Macro, stands out regarding the individual use of the feature categories. We realised, therefore, that the development of new linguistic features adds value in predicting the textual complexity of PorSimples texts.

\subsection{Complexity prediction of transcribed speech narratives of Adole-Sendo project}

\begin{figure}[!htb]
  \centering
  \includegraphics[scale=1.2]{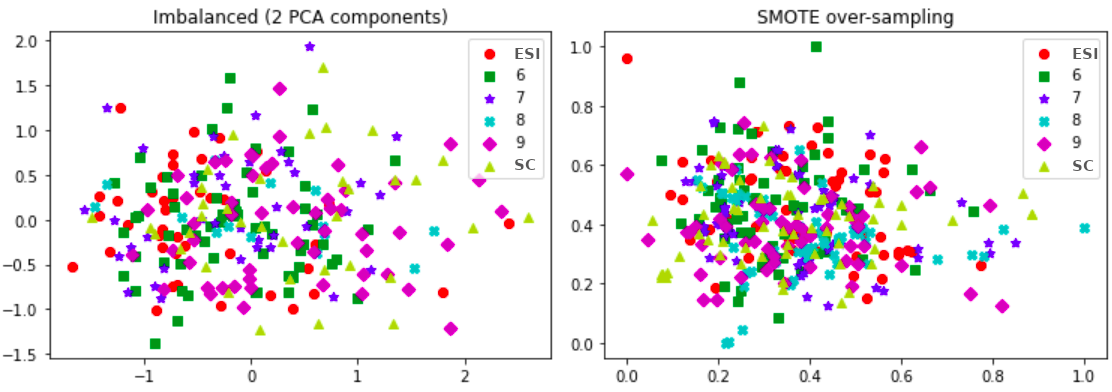}
  \caption{Adole-Sendo classes distribution plotted using PCA, before and after data-augmentation with SMOTE}
  \label{fig:adolesendo_smote}
\end{figure}

This experiment was performed to validate and exemplify the use of NILC-Metrix metrics applied to transcribed speech texts, using the 271 narratives of the Adole-Sendo corpus (see Section 4.1.3), grouped by grades and stages of education. As we are focusing on grades 6 to 9 of Elementary School II, our dataset comprises two new sets of narratives from the grades of Elementary School I, grouped, and of narratives from Secondary School. This division in six classes also helped to balance the samples: 4th and 5th grades were grouped in ESI (Elementary School I) totalling 43 texts; 6th grade has 70 texts, 7th grade has 43 texts, 8th grade has 15 texts, 9th grade has 59 texts and SC (Secondary School) has 41 texts. As can be seen in the Figure \ref{fig:adolesendo_smote}, the task is not trivial, as there is no clear separation between classes in two dimensions, for example.

We proceed the experiment with the normalisation of the 200 features using the MinMaxScaler which leaves all values between 0 and 1. Then, the ANOVA technique was used to select features \cite{brownlee2019}, reducing the number of relevant columns to 194 correlated with the classes; the top 20 more relevant features can be seen in Table \ref{tab:adolesendo-feats}.
10\% of each class of the dataset was also separated for validation (26 samples). For the remaining 245 samples, the classes were balanced using the SMOTE Over-Sampling \cite{Chawla2002} data-augmentation method. The result of this process can be seen in Figure \ref{fig:adolesendo_smote} where 63 samples were assigned per class.

\begin{table}[!htbp]
\centering
\begin{tabular}{llc}
\hline
\textbf{Name} & \textbf{Group} & \textbf{Weight} \\ \hline
cross\_entropy &	LSA-Semantic Cohesion	& 9.30 \\
prepositions\_per\_sentence &	Morphosyntactic Word Information	& 7.58 \\ 
first\_person\_pronouns &	Morphosyntactic Word Information	& 6.02 \\ 
long\_sentence\_ratio &	Text Easability Metrics	& 5.76 \\ 
content\_density &	Lexical Diversity	& 5.75 \\
verbs\_max &	Morphosyntactic Word Information	& 5.75 \\
prepositions\_per\_clause &	Morphosyntactic Word Information	& 5.65 \\
content\_words &	Morphosyntactic Word Information	& 5.56 \\ 
adverbs\_standard\_deviation &	Morphosyntactic Word Information	& 5.51 \\ 
function\_words & 	Morphosyntactic Word Information	& 5.47 \\ 
ratio\_function\_to\_content\_words & 	Morphosyntactic Word Information	& 5.29 \\ 
sentences\_with\_one\_clause & 	Syntactic Complexity	& 5.19 \\ 
adj\_arg\_ovl & 	Referential Cohesion	& 4.82 \\ 
dalechall\_adapted & 	Readability Formulas	& 4.79 \\ 
content\_word\_max & 	Lexical Diversity	& 4.65 \\ 
idade\_aquisicao\_mean & 	Psycholinguistic Measures	& 4.61 \\
arg\_ovl & 	Referential Cohesion	& 4.58 \\ 
non-inflected\_verbs	 & Morphosyntactic Word Information	& 4.50 \\ 
pronouns\_min & 	Morphosyntactic Word Information	& 4.45 \\ 
\hline
\end{tabular}
\caption{Top 20 features ordered by weight after selection with ANOVA technique on Adole-Sendo classification task}
\label{tab:adolesendo-feats}
\end{table}

\begin{figure}[!htb]
  \centering
  \includegraphics[scale=1.2]{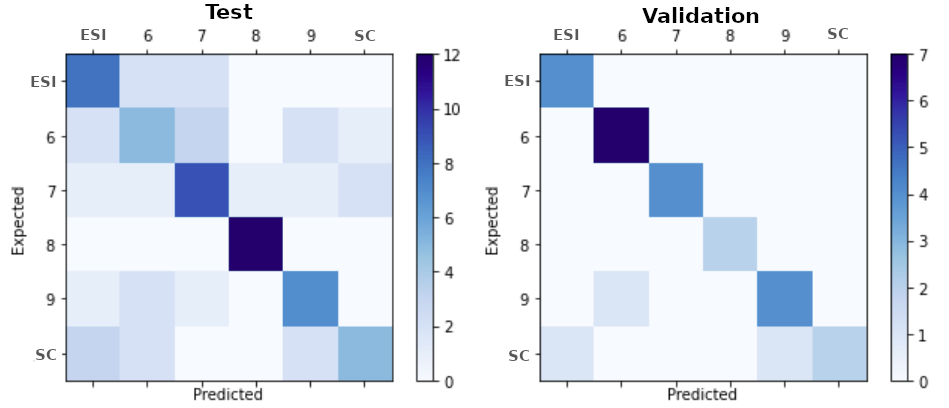}
  \caption{Confusion Matrix for test and validation samples from Adole-Sendo corpus}
  \label{fig:adolesendo_confusion}
\end{figure}

Five classification methods from the Scikit-Learn\footnote{\url{https://scikit-learn.org/stable/auto_examples/classification/plot_classifier_comparison.html}} library were chosen, using standard hyperparameters: a) Linear SVM with C = 0.025 ; b) SVB RBF with C = 1; c) Random Forest with max\_depth = 5; d) Neural Network MLP with 100 neurons in the hidden layer and e) Gaussian Naive Bayes. The best F1-Score method was the Neural Net with 0.62, but very close to SVM (Table \ref{tab:adolesendo-algoritmos}). The CV F-Score was calculated using 10-Fold Cross Validation and the Val F-Score was calculated from the prediction values in the validation dataset. Confusion matrices of test and validation data can be seen in Figure \ref{fig:adolesendo_confusion}.

\begin{table}[!htbp]
\centering
\begin{tabular}{lcc}
\hline
\textbf{Classifier} & \textbf{CV F-Score} & \textbf{Val. F-Score} \\ \hline
Linear SVM & 0.28 & 0.13 \\
RBF SVM & 0.61 & \textbf{0.88} \\ 
Random Forest & 0.39 & 0.68 \\ 
Neural Net & \textbf{0.62} & \textbf{0.88} \\ 
Naive Bayes & 0.38 & 0.48 \\ 
\hline
\end{tabular}
\caption{ML methods evaluated in the Adole-Sendo classification task, CV is 10-Fold Cross Validation and Val. is the result in reserved validation samples}
\label{tab:adolesendo-algoritmos}
\end{table}

\begin{table}[!htbp]
\centering
\begin{tabular}{lcc}
\hline
\textbf{Group} & \textbf{CV F1-Score} & \textbf{Std} \\ \hline
Lexical Diversity & 0.23 & 0.06 \\ 
Text Easability Metrics & 0.21 & 0.04 \\ 
Morphosyntactic  Word Information & 0.21 & 0.05 \\ 
Psycholinguistic Measures & 0.19 & 0.03 \\ 
Semantic Word Information & 0.18 & 0.05 \\ 
Descriptive Index & 0.15 & 0.03 \\ 
LSA-Semantic Cohesion & 0.15 & 0.04 \\ 
Temporal lexicon & 0.15 & 0.06 \\ 
Readability Formulas & 0.15 & 0.04 \\ 
Syntactic Complexity & 0.14 & 0.04 \\ 
Connectives & 0.14 & 0.03 \\ 
Word Frequency & 0.14 & 0.03 \\ 
Referential Cohesion & 0.12 & 0.02 \\ 
Syntactic Pattern Density & 0.11 & 0.02 \\ 
\hline
\end{tabular}
\caption{Evaluation of each group isolated features on Adole-Sendo classification task. CV F1-Score is the average of F1 with 10-Fold Cross Validation and Std is the standard deviation.}
\label{tab:adolesendo-groups}
\end{table}

Finally, the weight of each group of metrics was evaluated in the classification, using MLP Neural Net (the best method of the previous step). The set of metrics that performed best in isolation was \textbf{Lexical Diversity}, with 0.23 F1-Score, followed by \textbf{Text Easability Metrics} and \textbf{Morphosyntactic Word Information}. The complete list can be seen in the Table \ref{tab:adolesendo-groups}.

\section{Uses of NILC-Metrix Metrics}

In this section, we review 5	published studies in several research areas ---  Natural Language Processing, Neuropsychological Language Tests, Education, Language and Eye-tracking studies --- to illustrate	 the	 wide-ranging	use of sets of metrics available in NILC-Metrix.

\cite{santos-etal-2020-measuring} used 165 metrics 
of NILC-Metrix to evaluate their contribution to detect fake news for the BP language. The focus of the study was on 17 metrics of this large set, from 4 categories (Classic Readability Formulas, Referential Cohesion, Text Easability Metrics and Psycholinguistics), named as readability features by the authors.
The authors selected the following classic readability formulas: Flesch Index, Brunet Index, Honore Statistic,  
Dale Chall Formula, and Gunning Fog Index. From the set of 9 metrics of Referential Cohesion of NILC-Metrix, 7 of them were used: 4 metrics from the Psycholinguistic Measures and one from the set of Text Easability Metrics.
In their study the authors used an open access and balanced corpus called Fake.Br corpus\footnote{\url{https://github.com/roneysco/Fake.br-Corpus}}, with aligned texts totalling 3,600 false and 3,600 true news. SVM with the standard parameters of Scikit-learn\footnote{\url{https://scikit-learn.org/stable/index.html}} was used, along with traditional evaluation measures of precision, recall, F-measure and general
accuracy in a 5-fold cross-validation strategy.
The results of their study showed that readability features were relevant for detecting fake news in BP, achieving, alone, up to 92\% classification accuracy.

\cite{Aluisio_Propor16} evaluated classification and  regression methods to identify linguistic features for dementia diagnosis, focusing on Alzheimer Disease (AD) and Mild Cognitive Impairment (MCI), to distinguish them from Control Patients (CT). 
In their paper, a narrative language test was used based on sequenced pictures (Cinderella story) and features extracted from the resulting transcriptions, using the Coh-Metrix-Dementia tool.
It is important to note that the NILC-Metrix includes 18 metrics from Coh-Metrix-Dementia, 11 metrics from the LSA-Semantic Cohesion class, 4 from the Syntactic Complexity class, 2 Readability Formulas and one from the class Lexical Diversity. 
For the classification results, they obtained 0.82 F1-score in the experiment with three classes (AD, MCI and CT), and 0.90 for two classes (CT \textit{versus} (MCI+AD)), both using the CFS-selected features; for regression, they obtained 0.24 MAE for three classes, and 0.12 for two classes, both using all features available in the Coh-Metrix-Dementia tool. 

\cite{gazzolaStil2019} investigated the impact of textual genre in assessing text complexity in BP educational resources. Their final goal was to develop methods to assess the stage of education for the Open Educational Resources (OER) available on the platform MEC-RED (from the Brazilian Ministry of Education)\footnote{\url{https://plataformaintegrada.mec.gov.br/home}}.
For this purpose, a corpus with textbooks for Elementary School I, Elementary School II, Secondary School and Higher Education
was compiled. A set of 79 metrics from NILC-Metrix was selected, based on the study by \cite{graesser2011computational}.
Using those 79 metrics, they found correspondence which 53 metrics of Coh-Metrix, and grouped them into: \textit{Metrics Related to Words}, \textit{Related to Sentences} and \textit{Related to Connections between Sentences}. After selecting the features, 5 Machine Learning methods were tested: SVM, MLP, Logistic Regression and Random Forest from scikit learn\footnote{\url{https://scikit-learn.org/stable/}}. 
SVM performed better with 0.804 F-Measure, therefore it was used in an extrinsic evaluation with two sets of OER, reaching 0.518  F-Measure in the set with text genres similar from the training set (textbook corpus)  and 0.389 F-Measure for the animation/simulation and practical experiment resources, which are very common in the MEC-RED platform.

\cite{finatto-etal-2011-caracteristicas} evaluated the differences in text complexity of popular Brazilian newspapers (aimed at a public with
a lower education) with  traditional ones (aimed at more educated readers), using cohesion, syntax and vocabulary metrics, including ellipsis. In their contrastive analysis, the authors used 48 metrics from Coh-Metrix-Port and included 5 new ones related to the co-reference of ellipses, based on a corpus annotation. The annotation involved identifying ellipses of three types: nominal, verbal and sentential.
The study selected a balanced corpus of texts seeking the widest possible range of themes and editorials. They used 80
texts from the traditional Zero Hora newspaper from 2006 and 2007 and 80 texts from the popular Diário Gaucho from 2008\footnote{\url{https://gauchazh.clicrbs.com.br/}}. 
The authors found out that the most discriminative features between both newspapers were a set of 14 features grouped into 5 classes: Referential Cohesion, Word Frequency, Syntactic Complexity, Descriptive Index, Morphosyntactic Word Information,
extracted using Coh-Metrix-Port, but ellipsis did not have a distinctive role.

\cite{leal2019cluster} used NILC-Metrix metrics to propose a less subjective model for choosing texts and paragraphs for a project in the area of Psycholinguistics called RastrOS. In their study, the objective was to select 50 paragraphs with a wide range of language phenomena for RastrOS, a corpus with predictability norms and eye tracking data during silent reading of short paragraphs.
First, 58 metrics with great relevance to the task were manually selected (grouped into structural complexity, types of sentences, co-reference and morphosyntactics). Next, these metrics were extracted from all the paragraphs to help with grouping together texts with similar types of features by K-Means and Agglomerative Clustering methods. 
To assess the quality of the groups, the Elbow method, V-Measure and Silhouette techniques were used. After grouping, the paragraphs went through a human selection 
to find a few examples from each large text group.

\section{Concluding Remarks and Future Work}

The objective of this paper was to introduce and make the NILC-Metrix, a computational system comprising 200 metrics for BP, publicly available.  We presented the motivation for developing this large set of metrics and also illustrated the wide-ranging uses of NILC-Metrix published in studies of several research areas. We also presented three experiments based on corpora, using NILC-Metrix: an analysis of the differences between children's film subtitles and texts written for children, a new predictor of textual complexity for the PorSimples corpus, and a complexity prediction model for school grades, using transcripts of children's story narratives. For each case of study, we showed the robustness of NILC-Metrix, highlighting the importance of having a large number of metrics, that cover multiple linguistic aspects, available for textual analysis. 

Regarding future studies, we foresee two lines of research. The first one is related to implementing existing and new metrics and the NLP resources used for implementation.   For example, instead of using three parsers (LX-Parser\footnote{\url{http://lxcenter.di.fc.ul.pt/tools/pt/conteudo/LXParser.html}}, MaltParser\footnote{\url{http://www.maltparser.org/}} and Palavras) when implementing syntactic metrics, in the near future we will be able  to use robust parsing models for Portuguese, available in the POeTiSA project\footnote{\url{https://sites.google.com/icmc.usp.br/poetisa}}. As for new metrics, we also have a long list of suggestions. 
Idea Density is a metric that computes the number of propositions of a text, divided by its number of words; it was implemented in Coh-Metrix-Dementia \cite{CBMS_CMD_2015}, using a set of rules over dependency parsing\footnote{The metric uses a tool called IDD32 (Idea Density from Dependency Trees), which can extract propositions from well-written English and Portuguese texts, which is a drawback for its general use.}. 
Once a robust parsing model is made available, %by the POeTiSA project, 
the robustness of this metric can be evaluated and implemented in the NILC-Metrix.
\cite{flor-etal-2013-lexical} defined a metric called lexical
tightness that measures global cohesion of content words in a text.
According to the authors, this metric represents the degree to which a text tends to use words that are highly inter-associated in the language. This metric is a candidate to be evaluated and compared with the semantic cohesion metrics based on LSA, already implemented in the NILC-Metrix.  
\cite{Duran2007} evaluated temporal indices available in the Coh-Metrix in order to investigate temporal coherence. Six of the indices are available in the Coh-Metrix v3.0 and are related to the grammatical function (PoS, connectives that are already implemented in the NILC-Metrix, and temporal adverbial phrases); three other temporal indices were also proposed in their work. Temporal cohesion is also a candidate to be investigated and implemented for BP. Other sets of metrics related to causal and intentional cohesion available in Coh-Metrix should be studied and evaluated for BP. The second line of research is related to the validation of sets of metrics for several NLP tasks. We hope this paper can encourage researchers to work in this line of validating sets of metrics as we have made available both the access and code of the 200 metrics developed.

\section{Acknowledgments}
This work is part of the RastrOS Project supported by the São Paulo Research
Foundation (FAPESP—Regular Grant \#2019/09807-0).
The authors would like to thank all the members of the PorSimples project that provided the basis for building Coh-Metrix-Port and AIC metrics. We would also like to thank all the students who contributed (after PorSimples finished) to enlarging the set of metrics, revising it, applying it in various NLP tasks and, finally to making NILC-Metrix publicly available. 

\section{Declarations}

\textbf{Funding:} This research was supported by The São Paulo Research Foundation (FAPESP) (\textit{Fundação de Amparo à Pesquisa do Estado de São Paulo}, in Portuguese), Regular Grant \#2019/09807-0.

\textbf{Conflicts of interest/Competing interests:} The authors have no conflicts of interest to declare.

\textbf{Availability of data and material (data transparency):} Four datasets used in the applications of NILC-Metrix are available, in tsv format, in the file DATA at  \url{https://github.com/nilc-nlp/nilcmetrix}.

\textbf{Code availability (software application or custom code):} 
Source Code of NILC-Metrix is available at \url{https://github.com/nilc-nlp/nilcmetrix} under AGPLv3 license.

\textbf{Authors' contributions:}  

\textbf{Sidney Leal:} Conceptualisation,  Investigation, Methodology, Resources, Software Development, Validation, Writing – original paper; \textbf{Magali Duran:}
Conceptualisation, Data curation, Investigation,  Resources,  Writing – original paper; \textbf{Carolina Scarton:} Conceptualisation, Data curation,  Investigation, Methodology, Resources, Software Development, Validation, Writing – original paper; \textbf{Nathan Hartmann:} Conceptualisation, Data curation, Investigation, Methodology, Resources, Software Development, Validation, Writing – original paper; \textbf{Sandra Aluisio:} Conceptualisation, Data curation, Funding acquisition, Investigation, Methodology, Project administration, Resources, Supervision, Validation, Writing – original paper.

\bibliographystyle{acl}
\bibliography{main}

\end{document}